\documentclass{article}

\usepackage[main, final]{neurips_2026}

\usepackage[utf8]{inputenc} 
\usepackage[T1]{fontenc}    
\usepackage{url}            
\usepackage{booktabs}       
\usepackage{amsfonts}       
\usepackage{nicefrac}       
\usepackage{microtype}      

\usepackage[table]{xcolor}
\usepackage{xspace}
\usepackage{graphicx, amsmath, amssymb, caption, subcaption, multirow, overpic, textpos, pifont}
\usepackage{wrapfig}
\usepackage[american]{babel}
\usepackage[ruled]{algorithm2e}
\definecolor{citecolor}{HTML}{0071BC}
\definecolor{linkcolor}{HTML}{ED1C24}
\usepackage[pagebackref=false, breaklinks=true, colorlinks, citecolor=citecolor, linkcolor=linkcolor, bookmarks=false]{hyperref}
\newcommand{\etal}{\textit{et al.}}
\title{ProAlign: Progressive Positional and Prototype-Guided Alignment for Aerial-Ground Person Re-Identification}

\author{Qiao Li\textsuperscript{\rm 1}, Jing Chen\textsuperscript{\rm 1}\thanks{Corresponding Author: Jing Chen},
   Cong Wu\textsuperscript{\rm 1}, Kun He\textsuperscript{\rm 1} ,Ruiying Du\textsuperscript{\rm 1},Lei Tan\textsuperscript{\rm 2}\\
    \textsuperscript{\rm 1} School of Cyber Science and Engineering, Wuhan University\\
    \textsuperscript{\rm 2}National University of Singapore\\
  \texttt{liqiaoqiao233@whu.edu.cn, chenjing@whu.edu.cn}%
}

\begin{document}

\maketitle

\begin{abstract}
Aerial-Ground Person Re-Identification (AG-ReID) aims to match pedestrian images captured by unmanned aerial vehicles (UAVs) and ground-based surveillance cameras. The task remains highly challenging due to severe viewpoint discrepancies, frequent occlusions, and substantial domain gaps between aerial and ground imagery. Beyond these observable factors, existing methods often overlook two fundamental structural issues: cumulative positional drift during hierarchical feature extraction and semantic inconsistency across cross-view feature distributions.
To address these challenges, we propose ProAlign, a Progressive Positional and Prototype-Guided Alignment Network for AG-ReID. ProAlign comprises two key components: a Layer-wise Progressive Positional Embedding (LPPE) module and a View-Aware Prototype Contrastive Learning (PCL) module. LPPE performs adaptive positional calibration throughout transformer layers by employing conditional positional encoding in shallow layers to mitigate local spatial distortions, while introducing learnable static positional embeddings in deeper layers to reinforce global semantic priors. Meanwhile, PCL maintains view-specific aerial and ground prototypes for each identity and enforces cross-view semantic consistency via a dual-prototype contrastive objective.
Extensive experiments on CARGO and two real-world AG-ReID benchmarks demonstrate the effectiveness of ProAlign, achieving state-of-the-art overall retrieval performance on CARGO with improvements of 1.60\% in Rank-1 accuracy and 5.01\% in mAP over the strongest competing methods.
\end{abstract}

\section{Introduction}
Person re-identification (ReID) aims to match pedestrian images across distributed non-overlapping cameras and has attracted extensive attention due to its broad applications in intelligent surveillance systems. While conventional ReID has achieved remarkable progress~\cite{he2021transreid,li2023dc,tan2024partformer}, recent advances in unmanned aerial vehicles (UAVs) and low-altitude sensing platforms have stimulated growing interest in aerial-ground ReID (AG-ReID)~\cite{nguyen2023aerial,zhang2020person,zhang2024view}. Compared with traditional settings, AG-ReID introduces substantially larger viewpoint discrepancies and domain gaps between aerial and ground imagery, requiring models to establish robust cross-view identity correspondence under severe spatial and semantic variations.

Despite its practical importance, AG-ReID remains highly challenging and relatively underexplored. Unlike conventional ReID~\cite{tan2024rle,li2021diverse,wang2025towards}, aerial observations often exhibit extreme top-down viewpoints, severe occlusions, and incomplete body visibility, whereas ground images mainly contain frontal or profile appearances. Such discrepancies introduce significant geometric deformation and appearance variation across views. More importantly, these variations are not confined to the input level. As features propagate through deep Vision Transformer (ViT) architectures, early spatial deviations progressively accumulate and evolve into hierarchical positional drift, leading to unstable high-level representations. Meanwhile, the substantial appearance gap between aerial and ground domains produces systematic semantic inconsistency across feature distributions, making robust cross-view matching particularly difficult.

Recent AG-ReID methods attempt to alleviate these issues by learning view-invariant representations or introducing geometric and semantic alignment mechanisms~\cite{nguyen2024ag,zhang2024view,gsalign}. Although promising, most existing approaches primarily focus on shallow feature correction or sample-level alignment, while overlooking two fundamental structural limitations. First, positional misalignment propagates progressively across transformer hierarchies, where early spatial deviations become increasingly entangled in deeper semantic representations. Second, cross-view feature distributions remain globally inconsistent, whereas instance-level supervision alone is insufficient to preserve population-level semantic structure. Consequently, existing methods often struggle under large viewpoint discrepancies and severe visibility degradation.

\begin{figure}[t]
  \centering
  \includegraphics[width=\linewidth]{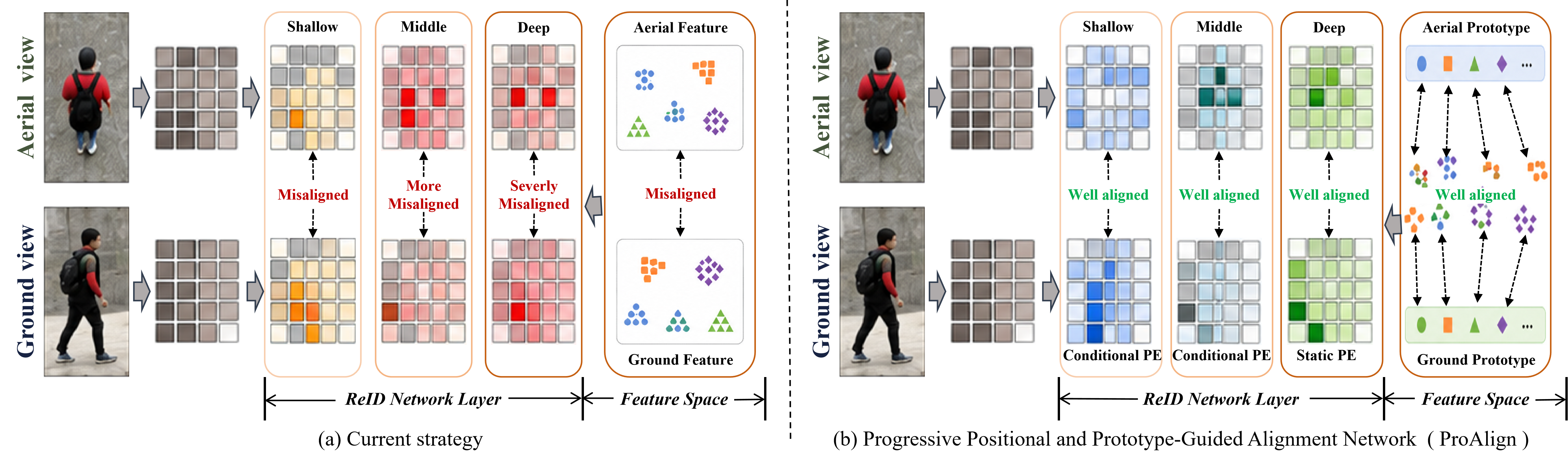}
   \caption{\textbf{Illustration of the motivation of ProAlign.} (a) Prior methods suffer from cumulative positional drift and cross-view distributional gaps. (b) ProAlign addresses these issues through layer-wise positional calibration (LPPE) and prototype-guided semantic alignment (PCL).}
   \vspace{-1em}
   \label{fig:intro}
\end{figure}

To address the above limitations, we propose ProAlign, a Progressive Positional and Prototype-Guided Alignment Network for AG-ReID, as illustrated in Figure~\ref{fig:intro}. The core idea of ProAlign is to jointly model hierarchical positional evolution and cross-view semantic structure within a unified framework.
Specifically, we introduce a Layer-wise Progressive Positional Embedding (LPPE) module to perform depth-adaptive positional calibration throughout transformer hierarchies. LPPE employs conditional positional encoding in shallow layers to mitigate local spatial distortions, while utilizing learnable static positional embeddings in deeper layers to reinforce stable global semantic priors. This progressive design enables the network to gradually correct spatial deviations without disrupting pretrained representations. Furthermore, we propose a View-Aware Prototype Contrastive Learning (PCL) module to explicitly align aerial and ground identity distributions at the identity level. PCL maintains view-specific prototypes for each identity and updates them online through quality-aware momentum aggregation, thereby reducing the influence of noisy or occluded samples. A dual-prototype contrastive objective further encourages features from different views to converge toward shared identity-aware semantic anchors, improving cross-view consistency and discriminative capability.

To sum up, the main contributions of this paper are as follows:
\begin{itemize}
\item We propose ProAlign, a unified framework for aerial-ground person re-identification that jointly addresses hierarchical positional drift and cross-view semantic inconsistency under extreme viewpoint variations.

\item We introduce a Layer-wise Progressive Positional Embedding (LPPE) module and a View-Aware Prototype Contrastive Learning (PCL) module. LPPE progressively calibrates spatial representations across transformer layers, while PCL performs prototype-guided population-level semantic alignment across views.

\item Extensive experiments on multiple AG-ReID benchmarks demonstrate that ProAlign achieves state-of-the-art overall performance and exhibits strong robustness under challenging aerial-ground matching scenarios.
\end{itemize}

\section{Related Work}
Building upon well-established ground surveillance infrastructure, ReID has achieved significant progress in a variety of settings, including occluded scenarios~\cite{wang2020high,li2021diverse,tan2024occluded,ye2024dynamic}, cross-modal scenarios~\cite{shao2023unified,qin2024noisy,zhao2024unifying,luo2025graph}, multi-spectral scenarios~\cite{wang2022interact,zhang2024magic,wang2025decoupled}, and cross-spectral scenarios~\cite{kim2023partmix,tan2023exploring,zhang2025adaptive,zhang2023diverse,ren2024implicit}. Despite these advances, most existing methods are designed under relatively stable camera viewpoints and limited cross-domain variation. With the rapid deployment of UAVs and low-altitude platforms, person ReID has been extended to aerial scenarios~\cite{nguyen2023ag}. Early works mainly explored aerial-aerial retrieval using datasets such as UAV-Human~\cite{li2021uav} and PRAI-1581~\cite{zhang2020person}. More recently, AG-ReID has attracted increasing attention due to its practical importance. This setting introduces severe viewpoint disparities and cross-view semantic inconsistencies, leading to substantial degradation in identity-preserving features~\cite{schumann2017person}. To address these challenges, recent methods explore view-invariant representation learning and cross-view alignment. Nguyen \etal~\cite{nguyen2023aerial,nguyen2024ag} propose attribute-guided multi-stream architectures. Zhang \etal~\cite{zhang2024view} introduce a View-Decoupled Transformer to separate view-specific and view-invariant features. GSAlign~\cite{gsalign} further models geometric deformation and visibility-aware alignment. 
Several recent works explicitly address geometric alignment: SeCap~\cite{wang2025secap} and GeoReID~\cite{hambarde2026rectifying} incorporate geometry-conditioned alignment, while GSAlign~\cite{gsalign} models geometric deformation and visibility-aware alignment via TPS structures. Furthermore, while memory-bank and prototype-based contrastive learning have been widely explored in conventional ReID, directly adapting these instance-level and global-memory mechanisms to AG-ReID often suffers from cross-view noise and domain gaps.
However, these methods primarily focus on input-level correction or instance-level alignment, while overlooking two structural issues: cumulative positional drift across transformer layers and global distributional inconsistency between aerial and ground representations. To address these limitations, we propose \textbf{ProAlign}, which jointly performs depth-adaptive positional calibration and prototype-guided cross-view alignment within a unified framework.

\section{Methodology}
\subsection{Overview}

\begin{figure}[t]
  \centering
  \includegraphics[width=\linewidth]{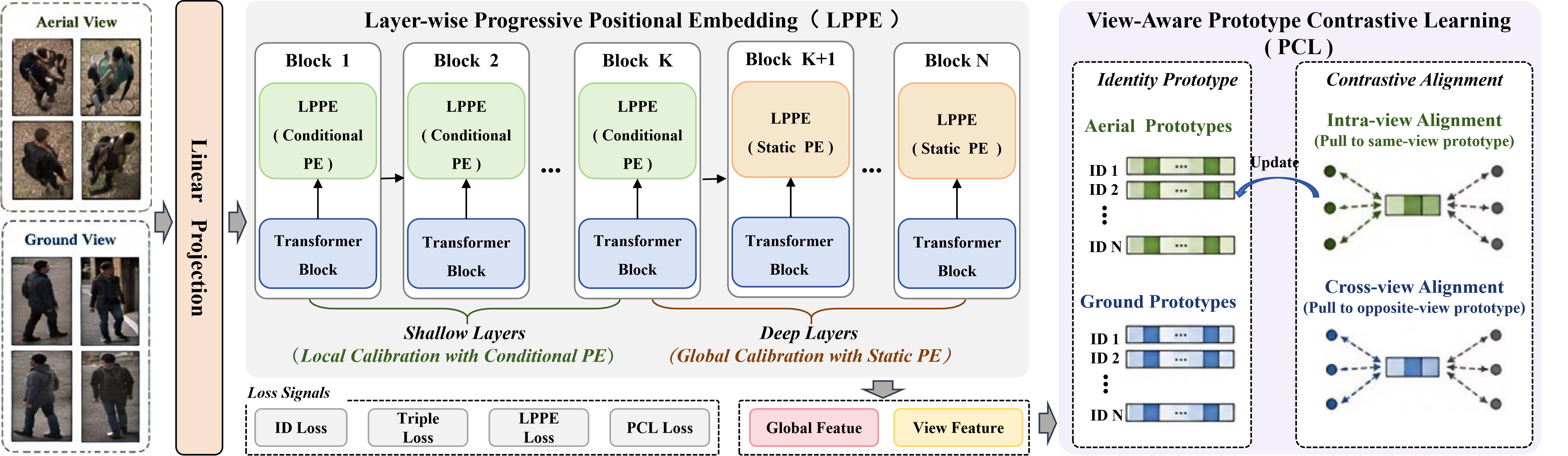}
   \caption{\textbf{Overview of the ProAlign architecture.} Given aerial or ground-view inputs, ProAlign first extracts patch tokens and processes them through transformer layers equipped with LPPE for depth-adaptive spatial correction. In parallel, the PCL module maintains view-specific prototypes to facilitate cross-view identity alignment.}
   \label{fig:model}\vspace{-1em}
\end{figure}

The proposed \textbf{Progressive Positional and Prototype-Guided Alignment Network (ProAlign)}, illustrated in Fig.~\ref{fig:model}, is built upon a ViT-Base backbone for AG-ReID. It aims to address hierarchical spatial degradation and cross-view semantic inconsistency under extreme viewpoint variations. Inspired by the success of the View-Decoupled Transformer (VDT)~\cite{zhang2024view}, we adopt the VDT architecture as the foundation of ProAlign, which effectively captures both global semantics and fine-grained structural cues. Following its design, we introduce an additional view token to model viewpoint-specific information. The token sequence during training is defined as:

\begin{equation}
\left[ \mathbf{X}_\mathrm{cls}, \mathbf{X}_\mathrm{v}, \mathbf{X}_\mathrm{img} \right] 
= \mathrm{VDT}\left( \left[ \mathbf{X}_\mathrm{cls}, \mathbf{X}_\mathrm{v}, \mathrm{tokenize}(\mathbf{X}_\mathrm{img}) \right] \right),
\end{equation}
where $\mathbf{X}_\mathrm{cls}$, $\mathbf{X}_\mathrm{v}$, and $\mathbf{X}_\mathrm{img}$ represent the class token, view token, and image tokens, respectively. 

To improve geometric robustness and mitigate cumulative positional drift, we integrate a \textbf{Layer-wise Progressive Positional Embedding (LPPE)} module into the transformer backbone. LPPE performs depth-adaptive positional calibration by applying conditional encodings in shallow layers and learnable static embeddings in deeper layers, progressively correcting pose-induced geometric distortions across network depth. In addition, we introduce a \textbf{View-Aware Prototype Contrastive Learning (PCL)} module during training to explicitly model identity distributions across aerial and ground views. PCL maintains quality-weighted prototypes that dynamically aggregate discriminative features, pulling samples toward consistent identity subspaces. This design effectively bridges the cross-view semantic gap while suppressing noise from background clutter and occlusions.

\subsection{Layer-wise Progressive Positional Embedding}

To address spatial misalignment in person re-identification, we design a Layer-wise Progressive Positional Embedding (LPPE) module that explicitly models geometric variations induced by viewpoint discrepancies. By injecting positional corrections at each transformer layer, LPPE prevents early spatial offsets from accumulating and degrading deep representations. As illustrated in Fig.~\ref{fig:model}, LPPE is integrated into each transformer block to enable hierarchical, depth-adaptive spatial alignment. Shallow layers primarily correct local spatial distortions, while deeper layers progressively reinforce global semantic positional priors.

The core idea of LPPE is to adopt depth-dependent positional encoding strategies across transformer layers. In early stages, where spatial deviations are largely local and texture-dependent, we employ Conditional Positional Encoding (CPE), which dynamically generates positional embeddings conditioned on local patch neighborhoods via depthwise convolution. Given a patch feature map $\mathbf{F}^{(l)} \in \mathbb{R}^{B \times C \times H \times W}$, the conditional positional correction $\mathbf{E}_c^{(l)}$ is formulated as:
\begin{equation}
    \mathbf{E}_c^{(l)} = \mathrm{DWConv}\!\left(\mathbf{F}^{(l)}\right),
\end{equation}
where $\mathrm{DWConv}(\cdot)$ denotes a $3 \times 3$ depthwise convolution with zero-initialized parameters. This design enables dynamic positional correction that adapts to input-specific local spatial misalignments.

In contrast, deeper layers in ViT encode highly abstract and globally entangled representations. To stabilize these features, we introduce learnable static positional embeddings. Unlike conditional encodings, we maintain a set of independent, layer-specific embeddings $\mathbf{E}_s^{(l)} \in \mathbb{R}^{1 \times N \times C}$, which serve as global semantic anchors to guide high-level representations toward a canonical spatial configuration.

To preserve pretrained representations and avoid disrupting learned semantics, we further introduce a learnable gating mechanism. The positional correction is applied only to patch tokens, while class and view tokens remain unchanged. The calibrated patch features $\tilde{\mathbf{F}}^{(l)}$ at layer $l$ are obtained via gated residual fusion:
\begin{equation}
    \tilde{\mathbf{F}}^{(l)} = \mathbf{F}^{(l)} + g_l \cdot \mathbf{E}^{(l)},
\end{equation}
where $\mathbf{E}^{(l)}$ denotes the positional embedding at layer $l$, selected from either $\mathbf{E}_c^{(l)}$ or $\mathbf{E}_s^{(l)}$, and $g_l$ is a learnable scalar initialized to zero. This gating mechanism gradually activates LPPE during training, enabling a smooth introduction of spatial corrections while maintaining stability of pretrained representations under large cross-view variations.

\subsection{View-Aware Prototype Contrastive Learning}
In AG-ReID, severe viewpoint variations introduce a substantial distributional gap between aerial and ground representations of the same identity. Direct sample-to-sample alignment is often unreliable due to occlusions and background noise. To address this issue, we propose View-Aware Prototype Contrastive Learning (PCL), which explicitly models population-level identity distributions to bridge the cross-view semantic gap. PCL maintains a momentum-updated memory bank that anchors each identity in a shared prototype space, thereby improving global feature consistency while reducing the influence of noisy samples.

Specifically, we construct a view-conditioned prototype memory bank during training. For each identity $c$, we maintain two view-specific prototypes: an aerial prototype $\mathbf{p}_c^a \in \mathbb{R}^D$ and a ground prototype $\mathbf{p}_c^g \in \mathbb{R}^D$. To enhance robustness, we introduce a quality-aware momentum update strategy that prevents occluded or low-quality samples from degrading the prototype representations. The quality weight $w_i$ for the $i$-th normalized feature $\hat{\mathbf{f}}_i = \mathbf{f}_i/\|\mathbf{f}_i\|_2$ of class $c$ in view $v$ is defined as its cosine similarity with the corresponding prototype:

\begin{equation}
    w_i = \max\left(0,\; \hat{\mathbf{f}}_i \cdot \mathbf{p}_c^v \right).
\end{equation}

For each newly observed identity-view pair, the prototype
is initialized using the normalized mean of its samples.
For subsequent updates, nonnegative cosine similarities
are used as quality weights to reduce the influence of
potentially noisy features. If the sum of weights is
negligible, uniform averaging is adopted to ensure a
stable prototype update.

The prototypes are then updated using a momentum-based strategy. Let $\overline{\mathbf{f}}_c^v$ denote the quality-weighted mean of valid samples for class $c$ and view $v$ within the current mini-batch. The update rule is defined as:
\begin{equation}
    \mathbf{p}_c^v \leftarrow \mathrm{Norm} \left( (1 - m) \cdot \mathbf{p}_c^v + m \cdot \overline{\mathbf{f}}_c^v \right),
\end{equation}
where $m \in [0, 1)$ is the momentum coefficient and $\mathrm{Norm}$ denotes $\ell_2$ normalization. For newly initialized identities, the prototype is directly set to the normalized sample mean without momentum. This weighted update scheme ensures that prototypes are dominated by discriminative and high-visibility features, improving robustness to occlusion and background noise.

During training, these prototypes serve as anchors for contrastive learning. Each sample is encouraged to align with its corresponding prototypes in both intra-view and cross-view spaces, thereby promoting viewpoint-invariant representation learning. Importantly, the prototype memory bank is used only during training to structure the feature space, and introduces no additional computational overhead during inference.

\subsection{Loss Function and Optimization}

Our model is trained with a composite objective that jointly optimizes identity discrimination, hierarchical spatial stability, and cross-view semantic alignment. In addition to the standard ID classification loss $\mathcal{L}_\mathrm{id}$ and Triplet loss $\mathcal{L}_\mathrm{tri}$ commonly used in ReID, we introduce a smoothness regularization loss $\mathcal{L}_\mathrm{sm}$ and a prototype contrastive loss $\mathcal{L}_\mathrm{pcl}$.

\textbf{Smoothness Regularization Loss.} To stabilize the static positional embeddings in deeper LPPE layers, we introduce a smoothness constraint that discourages abrupt variations between consecutive layers. This encourages gradual refinement of global positional representations. The loss is defined as the mean squared error between adjacent static embeddings:
\begin{equation}
\mathcal{L}_\mathrm{sm}
=
\frac{1}{|\mathcal{L}_s|-1}
\sum_{l\in\mathcal{L}_s\setminus
\{\max(\mathcal{L}_s)\}}
\left\|
\mathbf{E}_s^{(l+1)}
-
\mathbf{E}_s^{(l)}
\right\|_F^2.
\end{equation}
where $\mathcal{L}_s$ denotes the set of layers equipped with static positional embeddings. This constraint promotes smooth hierarchical evolution of global spatial priors and avoids unstable layer-wise positional oscillations.

\textbf{Prototype Contrastive Loss.} To enforce cross-view semantic consistency at the population level, we formulate a dual-prototype InfoNCE objective $\mathcal{L}_\mathrm{pcl}$, consisting of intra-view and cross-view alignment terms. 
Given a normalized feature $\hat{\mathbf{f}}_i$
with identity label $c$ from view
$v\in\{0,1\}$, where 0 and 1 denote
ground and aerial views, respectively, the model contrasts it against all identity prototypes stored in the memory bank. The InfoNCE loss for target view $t$ is defined as:
\begin{equation}
    \mathcal{L}_\mathrm{nce}(\hat{\mathbf{f}}_i, t) = -\log \frac{\exp(\hat{\mathbf{f}}_i \cdot \mathbf{p}_c^t / \tau)}{\sum_{k=1}^{C} \exp(\hat{\mathbf{f}}_i \cdot \mathbf{p}_k^t / \tau)},
\end{equation}
where $\tau$ is the temperature and $C$ is the number of identities in the memory bank.

The intra-view term enforces consistency within the same viewpoint ($t=v$), while the cross-view term aligns features across different viewpoints ($t=1-v$). The overall loss is formulated as:

\begin{equation}
\mathcal{L}_\mathrm{pcl}
= \frac{1}{N}\sum_{i=1}^{N}
\Big(
\mathcal{L}_\mathrm{nce}(\hat{\mathbf{f}}_i,v_i)
+
\gamma\,
\mathcal{L}_\mathrm{nce}(\hat{\mathbf{f}}_i,1-v_i)
\Big).
\end{equation}

where $N$ denotes the batch size and $\gamma$ controls the strength of cross-view alignment. This objective explicitly encourages viewpoint-invariant representation learning by aligning samples to shared identity prototypes across views.

\textbf{Overall Loss.} The model is trained end-to-end by minimizing a comprehensive loss function that aggregates the aforementioned components:

\begin{equation}
    \mathcal{L}_\mathrm{total} = ( \mathcal{L}_\mathrm{id} + \mathcal{L}_\mathrm{tri} ) + \alpha\, \mathcal{L}_\mathrm{sm} + \beta\, \mathcal{L}_\mathrm{pcl}.
\end{equation}
where $\alpha$ and $\beta$ balance the contributions of spatial regularization and prototype-level alignment. This joint optimization yields representations that are both spatially stable and robust to cross-view semantic discrepancies.

\section{Experiments}
\subsection{Implementation}

\textbf{Datasets and Evaluation Protocols.}
We evaluate ProAlign on the synthetic CARGO dataset~\citep{zhang2024view}
and two real-world datasets, AG-ReID~\citep{nguyen2023aerial}
and AG-ReID v2~\citep{nguyen2024ag}.
CARGO provides diverse aerial and ground views under challenging
viewpoint variations, resolution changes, and occlusions.
We follow the official data splits and evaluation protocols
for all datasets. On CARGO, we evaluate four retrieval settings:
\textit{ALL}, \textit{G$\leftrightarrow$G},
\textit{A$\leftrightarrow$A}, and
\textit{A$\leftrightarrow$G}.
We report Rank-1, mAP, and mINP~\citep{ye2021deep},
except for AG-ReID v2, where mINP is not reported
following prior works.
All ablation and parameter studies are conducted on CARGO.

\textbf{Implementation Details.} All settings follow VDT~\citep{zhang2024view} for fair comparison. We adopt an ImageNet-21K pretrained ViT-Base backbone with $256 \times 128$ inputs and horizontal flipping augmentation. The model is trained for 120 epochs using AdamW (weight decay 0.05) with an initial learning rate of $3.5 \times 10^{-4}$, a 20-epoch linear warm-up, and cosine decay. Each mini-batch contains $P=64$ identities with $K=4$ instances each. We set $\alpha=\beta=1.0$, $\gamma=0.5$, $m=0.2$, and apply CPE to the first 6 layers and static embeddings to the remaining 6. All experiments are conducted on an Ubuntu 20.04 system with an Intel Xeon Gold 6133 CPU @ 2.50GHz and an NVIDIA GeForce RTX 4090 GPU. The implementation is based on Python 3.8 and PyTorch 1.12.0 with CUDA 12.0. 

\subsection[Comparison with State-of-the-Arts]{Comparison with State-of-the-Arts\footnotemark}
\footnotetext{Extended comparisons on the AG-ReID and AG-ReID v2 datasets (Sec.~\ref{sec:agreid}) are provided in the supplementary material.}

We compare ProAlign with a range of state-of-the-art person ReID methods on the CARGO dataset, including CNN-based approaches (e.g., PCB~\citep{sun2019learning}, BoT~\citep{luo2019bag}, MGN~\citep{wang2018learning}), Transformer-based models (e.g., ViT~\citep{dosovitskiy2020image}, VDT~\citep{zhang2024view}, VIF~\citep{Khalid_2025_ICCV}), and the alignment-based GSAlign~\citep{gsalign}. As shown in Table~\ref{performance-cargo}, ProAlign achieves the best Rank-1, mAP and mINP under the ALL, G$\leftrightarrow$G and A$\leftrightarrow$A protocols. Under ALL, it obtains 68.27\% Rank-1, 62.47\% mAP and 51.39\% mINP, outperforming GSAlign by +1.60\%, +6.12\% and +9.97\%, respectively. On the challenging A$\leftrightarrow$G protocol, ProAlign improves over its VDT baseline by +12.52\% Rank-1, +15.69\% mAP and +19.91\% mINP, and achieves the best mINP (49.86\%). GSAlign obtains higher A$\leftrightarrow$G Rank-1 and mAP through explicit TPS-based geometric warping, but at the cost of 25.7M additional parameters (Table~\ref{tab:complexity}), whereas ProAlign adds only 1.01M and requires no geometric transformation. Notably, ProAlign also improves single-view protocols, indicating that the proposed alignment does not sacrifice intra-view retrieval.

\begin{table*}[t]
	\centering
	\caption{\textbf{Performance comparison of the mainstream methods under four settings of the CARGO dataset}. ``ALL'' denotes the overall retrieval performance of each method. ``G$\leftrightarrow$G'', ``A$\leftrightarrow$A'', and ``A$\leftrightarrow$G'' represent the performance of each model in several specific retrieval patterns. Rank1, mAP, and mINP are reported (\%). The best performance is shown in \textbf{bold}.}
	\label{performance-cargo}
	\renewcommand{\arraystretch}{1.2}
	\resizebox{\textwidth}{!}{%
		\begin{tabular}{c|ccc|ccc|ccc|ccc}
			\hline
			\multirow{2}{*}{Method} & \multicolumn{3}{c|}{Protocol 1: ALL} & \multicolumn{3}{c|}{Protocol 2: G$\leftrightarrow$G} & \multicolumn{3}{c|}{Protocol 3: A$\leftrightarrow$A} & \multicolumn{3}{c}{Protocol 4: A$\leftrightarrow$G} \\ \cline{2-13}
			& Rank1 & mAP & mINP & Rank1 & mAP & mINP & Rank1 & mAP & mINP & Rank1 & mAP & mINP \\ \hline
			SBS \cite{he2023fastreid} & 50.32 & 43.09 & 29.76 & 72.31 & 62.99 & 48.24 & 67.50 & 49.73 & 29.32 & 31.25 & 29.00 & 18.71 \\
			PCB \cite{sun2019learning} & 51.00 & 44.50 & 32.20 & 74.10 & 67.60 & 55.10 & 55.00 & 44.60 & 27.00 & 34.40 & 30.40 & 20.10 \\
			BoT \cite{luo2019bag} & 54.81 & 46.49 & 32.40 & 77.68 & 66.47 & 51.34 & 65.00 & 49.79 & 29.82 & 36.25 & 32.56 & 21.46 \\
			MGN \cite{wang2018learning} & 54.81 & 49.08 & 36.52 & 83.93 & 71.05 & 55.20 & 65.00 & 52.96 & 36.78 & 31.87 & 33.47 & 24.64 \\
			VV \cite{kuma2019vehicle, kumar2020strong} & 45.83 & 38.84 & 39.57 & 72.31 & 62.99 & 48.24 & 67.50 & 49.73 & 29.32 & 31.25 & 29.00 & 18.71 \\
			AGW \cite{ye2021deep} & 60.26 & 53.44 & 40.22 & 81.25 & 71.66 & 58.09 & 67.50 & 56.48 & 40.40 & 43.57 & 40.90 & 29.39 \\
			ViT \cite{dosovitskiy2020image} & 61.54 & 53.54 & 39.61 & 82.14 & 71.34 & 57.55 & 80.00 & 64.67 & 47.07 & 43.13 & 40.11 & 28.20 \\
            GSAlign \cite{gsalign} & 66.67 & 56.35 & 41.42 & 83.04 & 71.57 & 57.68 & 82.50 & 67.70 & 51.80 & \textbf{65.96} & \textbf{60.50} & 49.32 \\ 
            VDT~\cite{zhang2024view} & 64.10 & 55.20 & 41.13 & 82.14 & 71.59 & 58.39 & 82.50 & 66.83 & 50.22 & 48.12 & 42.75 & 29.95 \\ 
            VIF~\cite{Khalid_2025_ICCV}& 65.71	& 57.46	& 44.12	& 83.93	& 74.19	& 62.30	& 82.50	& 66.98	& 51.44	& 51.25	& 44.55	& 31.20\\
\hline
\rowcolor[gray]{0.92}\textbf{ProAlign} & \textbf{68.27} & \textbf{62.47} & \textbf{51.39} & \textbf{84.82} & \textbf{80.22} & \textbf{72.51} & \textbf{85.00} & \textbf{70.05} & \textbf{54.66} & 60.64 & 58.44 & \textbf{49.86} \\ \hline
		\end{tabular}
	}
\end{table*}

Overall, ProAlign consistently improves retrieval performance and remains robust across diverse aerial-ground scenarios. We observe stable gains on both the CARGO dataset and real-world benchmarks. Results on AG-ReID and AG-ReID v2 are reported in Sec.~\ref{sec:agreid} of the supplementary material, where ProAlign also achieves consistent improvements across settings. These results indicate that the proposed positional calibration and prototype alignment mechanisms are effective and generalize across different domains.

\subsection{Ablation Study}
\begin{table*}[!t]
	\centering
	\caption{\textbf{Ablation study of the different components in ProAlign on the CARGO dataset.} ``LPPE'' denotes the Layer-wise Positional Embedding module. ``PCL'' refers to the Prototype Contrastive Learning module. The best performance per metric is shown in \textbf{bold}.}
	\label{Tbl:ablation}
	\renewcommand{\arraystretch}{1.2}
	\resizebox{\textwidth}{!}{%
		\begin{tabular}{l|ccc|ccc|ccc|ccc}
			\hline
			\multirow{2}{*}{\textbf{Setting}} & \multicolumn{3}{c|}{Protocol 1: ALL} & \multicolumn{3}{c|}{Protocol 2: G$\leftrightarrow$G} & \multicolumn{3}{c|}{Protocol 3: A$\leftrightarrow$A} & \multicolumn{3}{c}{Protocol 4: A$\leftrightarrow$G} \\ \cline{2-13} 
			& Rank1 & mAP & mINP & Rank1 & mAP & mINP & Rank1 & mAP & mINP & Rank1 & mAP & mINP \\ 
            \hline
			Baseline & 64.10 & 55.20 & 41.13 & 82.14 & 71.59 & 58.39 & 82.50 & 66.83 & 50.22 & 48.12 & 42.75 & 29.95  \\
			Baseline + LPPE & 60.58	&54.92	&43.13	&79.46	&74.10	&65.12	&75.00 	&63.62	&47.28 &53.19	&52.63	&44.73\\
            Baseline + PCL  & 65.38 & 58.55 & 46.49 & 79.46 & 73.99 & 64.52 & 82.50 & 67.66 & 51.20 & 58.51 & 57.65 & 49.68 \\
            \hline
			\rowcolor[gray]{0.92} Baseline + LPPE + PCL & \textbf{68.27} & \textbf{62.47} & \textbf{51.39} & \textbf{84.82} & \textbf{80.22} & \textbf{72.51} & \textbf{85.00} & \textbf{70.05} & \textbf{54.66} & \textbf{60.64} & \textbf{58.44} & \textbf{49.86} \\
            \hline
		\end{tabular}
	}\vspace{-1em}
\end{table*}

To evaluate the contribution of each component in ProAlign, we conduct an ablation study on all four protocols of the CARGO dataset. As shown in Table~\ref{Tbl:ablation}, we start from the VDT baseline~\citep{zhang2024view}. \textbf{LPPE alone} yields large gains on the cross-view A$\leftrightarrow$G protocol (+5.07 Rank-1, +9.88 mAP, +14.78 mINP) and improves G$\leftrightarrow$G mAP/mINP, but it reduces Rank-1 on ALL and the intra-view protocols, as well as A$\leftrightarrow$A mAP/mINP. We attribute this to the fact that, without identity-level cross-view supervision, the newly introduced positional parameters are optimized mainly toward viewpoint-invariant spatial layouts, which slightly weakens view-specific cues useful for intra-view matching. Note that the A$\leftrightarrow$A protocol contains only 40 queries, so each query changes Rank-1 by 2.5\% (the $-7.50$ drop corresponds to three queries). \textbf{PCL alone} improves ALL (+1.28/+3.35/+5.36) and substantially improves A$\leftrightarrow$G (+10.39/+14.90/+19.73), showing that prototype alignment is independently effective, although it slightly lowers G$\leftrightarrow$G Rank-1. \textbf{Combining both} gives the best result on every metric of every protocol, including all intra-view metrics where LPPE alone falls behind (e.g., A$\leftrightarrow$A Rank-1 from 75.00 to 85.00). This indicates that the two modules are complementary: PCL anchors the identity semantics so that LPPE's spatial calibration no longer trades off intra-view discrimination.

\subsection[Discussions]{Discussions\footnotemark}
\footnotetext{A Visualization Analysis of LPPE (Sec.~\ref{sup:vis}) is provided in the supplementary material.}

\begin{table*}[!t]
	\centering
	\caption{\textbf{Performance comparison under different depths of CPE in LPPE.} Results are reported under four matching protocols on the CARGO dataset. The best values per column are shown in \textbf{bold}.}
	\label{Tbl:cpe_ratio}
	\renewcommand{\arraystretch}{1.2}
	\resizebox{\textwidth}{!}{%
		\begin{tabular}{c|ccc|ccc|ccc|ccc}
			\hline
			\multirow{2}{*}{\textbf{Setting}} & \multicolumn{3}{c|}{Protocol 1: ALL} & \multicolumn{3}{c|}{Protocol 2: G$\leftrightarrow$G} & \multicolumn{3}{c|}{Protocol 3: A$\leftrightarrow$A} & \multicolumn{3}{c}{Protocol 4: A$\leftrightarrow$G} \\ \cline{2-13} 
			& Rank1 & mAP & mINP & Rank1 & mAP & mINP & Rank1 & mAP & mINP & Rank1 & mAP & mINP \\ 
            \hline
            \multicolumn{13}{l}{\textbf{Depths of CPE}}\\
            \hline
           First 3 layers	&\textbf{68.59}	&60.97	&48.7	&\textbf{85.71}	&77.74	&67.86	&80.00	&68.32	&53.99	&60.64	&60.31	&52.22\\
           First 6 layers	&68.27	&\textbf{62.47}	&\textbf{51.39}	&84.82	&\textbf{80.22}	&\textbf{72.51}	&\textbf{85.00}	&\textbf{70.05}	&\textbf{54.66}	&60.64	&58.44	&49.86\\
           First 9 layers	&66.35	&60.40	&48.43	&81.25	&76.22	&67.17	&82.50	&68.73	&53.71	&\textbf{62.77}	&\textbf{61.66}	&\textbf{53.87}\\
           All layers	    &67.63	&58.87	&45.41	&83.93	&74.92	&64.11	&80.00	&67.58 	&52.16	 &61.70	&59.60	&50.12\\ 
            \hline
		\end{tabular}%
	}
\end{table*}

\noindent\textbf{Depth of Conditional Positional Encoding in LPPE.}
We vary the number of transformer layers equipped with CPE while using static positional embeddings for the remaining layers (Table~\ref{Tbl:cpe_ratio}). We select the configuration by a predefined criterion, the best ALL-protocol mAP/mINP, rather than per-protocol tuning. Applying CPE to the first six layers achieves the best mAP/mINP on ALL, G$\leftrightarrow$G and A$\leftrightarrow$A. Using CPE only in the first three layers yields the highest ALL and G$\leftrightarrow$G Rank-1 but lower mAP/mINP, while deeper CPE (first nine layers) gives the best A$\leftrightarrow$G results. Applying CPE to all layers yields the lowest ALL mINP (45.41\%), suggesting that input-conditioned encoding throughout the network weakens overall ranking quality. These results reveal a trade-off between intra-view and cross-view retrieval across CPE depths; the first-six configuration offers the best overall balance. Rank-1 differences are partly within the granularity of CARGO's small query sets (one query changes A$\leftrightarrow$A Rank-1 by 2.5\%). Notably, the full model with every CPE depth outperforms the VDT baseline on ALL mAP/mINP (Table~\ref{Tbl:ablation}), and all variants preserve spatial focus in the middle layers (Table~\ref{Tbl:entropy}).

\begin{table*}[!t]
	\centering
	\caption{\textbf{Impact of the cross-view alignment weight $\gamma$ in PCL.} Results are reported under four matching protocols on the CARGO dataset. The best values per column are shown in \textbf{bold}.}
	\label{Tbl:gamma_weight}
	\renewcommand{\arraystretch}{1.2}
	\resizebox{\textwidth}{!}{%
		\begin{tabular}{c|ccc|ccc|ccc|ccc}
			\hline
			\multirow{2}{*}{\textbf{Setting}} & \multicolumn{3}{c|}{Protocol 1: ALL} & \multicolumn{3}{c|}{Protocol 2: G$\leftrightarrow$G} & \multicolumn{3}{c|}{Protocol 3: A$\leftrightarrow$A} & \multicolumn{3}{c}{Protocol 4: A$\leftrightarrow$G} \\ \cline{2-13} 
			& Rank1 & mAP & mINP & Rank1 & mAP & mINP & Rank1 & mAP & mINP & Rank1 & mAP & mINP \\ 
            \hline
            $\gamma=0$    & 66.67 & 57.24 & 42.85 & 82.14 & 73.79 & 62.44 & 80.00 & 64.44 & 46.03 & \textbf{63.83} & 57.99 & 46.82 \\ 
            $\gamma=0.25$ & 67.31 & 58.54 & 45.12 & 82.14 & 75.25 & 65.02 & \textbf{85.00} & 68.98 & 51.75 & 62.77 & \textbf{60.54} & \textbf{51.29} \\ 
             $\gamma=0.5$  & \textbf{68.27} & \textbf{62.47} & \textbf{51.39} & \textbf{84.82} & \textbf{80.22} & \textbf{72.51} & \textbf{85.00} & \textbf{70.05} & \textbf{54.66} & 60.64 & 58.44 & 49.86 \\ 
            $\gamma=0.75$ & 66.67 & 58.54 & 44.67 & 82.14 & 73.25 & 61.65 & 82.50 & 68.39 & 53.43 & 59.57 & 58.76 & 49.85 \\ 
            \hline
		\end{tabular}%
	}
\end{table*}

\begin{table*}[!t]
	\centering
	\caption{\textbf{Comparison between different variants of prototype.} Results are reported under four matching protocols on the CARGO dataset. The best values per column are shown in \textbf{bold}.}
	\label{Tbl:vproto}
	\renewcommand{\arraystretch}{1.2}
	\resizebox{\textwidth}{!}{%
		\begin{tabular}{c|ccc|ccc|ccc|ccc}
			\hline
			\multirow{2}{*}{\textbf{Setting}} & \multicolumn{3}{c|}{Protocol 1: ALL} & \multicolumn{3}{c|}{Protocol 2: G$\leftrightarrow$G} & \multicolumn{3}{c|}{Protocol 3: A$\leftrightarrow$A} & \multicolumn{3}{c}{Protocol 4: A$\leftrightarrow$G} \\ \cline{2-13} 
			& Rank1 & mAP & mINP & Rank1 & mAP & mINP & Rank1 & mAP & mINP & Rank1 & mAP & mINP \\ 
            \hline
            \multicolumn{13}{l}{\textbf{Different variants of prototype}}\\
            \hline
            Inner-Batch & 62.82  & 56.12 & 43.55 & 77.68 & 72.57 & 62.64 & 77.50 & 64.44 & 48.94 & 60.64 & \textbf{58.50} & 49.49 \\ 
            Global Memory & 61.86 & 55.40 & 42.71 & 77.68 & 71.83 & 61.10 & 77.50 & 64.53 & 48.89 & 56.38 & 55.80 & 47.80 \\ 
            Classifier Weights & 62.82 & 56.30 & 43.98 & 81.25 & 75.06 & 65.08 & 75.00 & 63.90 & 47.79 & 53.19 & 53.52 & 45.54 \\ 
            \hline
            \rowcolor[gray]{0.92} View-Aware PCL& \textbf{68.27} & \textbf{62.47} & \textbf{51.39} & \textbf{84.82} & \textbf{80.22} & \textbf{72.51} & \textbf{85.00} & \textbf{70.05} & \textbf{54.66} & \textbf{60.64} & 58.44 & \textbf{49.86} \\ 
            \hline
		\end{tabular}%
	}\vspace{-1.5em}
\end{table*}

\noindent\textbf{Impact of cross-view alignment weight in PCL.}
The weight $\gamma$ controls the trade-off between intra-view discrimination and cross-view alignment (Table~\ref{Tbl:gamma_weight}). Without cross-view supervision ($\gamma=0$), the model attains the highest A$\leftrightarrow$G Rank-1 but the lowest ALL mAP/mINP (57.24\%/42.85\%) and clearly lower A$\leftrightarrow$G mAP/mINP, indicating that it retrieves the easiest cross-view matches but ranks the remaining true matches poorly. Increasing $\gamma$ to 0.5 steadily improves ALL, G$\leftrightarrow$G and A$\leftrightarrow$A, reaching 62.47\% mAP and 51.39\% mINP on ALL, while A$\leftrightarrow$G Rank-1 decreases moderately. Further increasing $\gamma$ to 0.75 degrades ALL and G$\leftrightarrow$G performance without improving A$\leftrightarrow$G, suggesting that overly strong alignment suppresses discriminative identity cues. We therefore adopt $\gamma=0.5$ following the same ALL-based selection criterion.

\noindent\textbf{Comparison of prototype variants.}
We evaluate different prototype construction strategies, including Inner-Batch, Global Memory, Classifier Weights, and our View-Aware PCL. As shown in Table~\ref{Tbl:vproto}, clear performance gaps appear across all four protocols. Compared with Inner-Batch prototypes, which achieve 62.82\% Rank-1 and 56.12\% mAP under the ALL protocol, Global Memory slightly reduces performance to 61.86\% Rank-1 and 55.40\% mAP, indicating that global averaging introduces noise from occluded or low-quality samples. Classifier Weights achieve baseline performance comparable to Inner-Batch under the ALL protocol (62.82\% Rank-1), but generalize poorly to cross-view retrieval, dropping to 53.19\% Rank-1 on A$\leftrightarrow$G. This suggests that classification-oriented prototypes are suboptimal for cross-view discrimination. In contrast, our View-Aware PCL achieves the best results on ALL, G$\leftrightarrow$G and A$\leftrightarrow$A, improving ALL Rank-1 by +5.45\% over Inner-Batch and Classifier Weights, and G$\leftrightarrow$G mINP by +7.43\% over the strongest variant. On A$\leftrightarrow$G, it matches Inner-Batch in Rank-1 and achieves the best mINP, while outperforming Global Memory and Classifier Weights by +4.26\% and +7.45\% Rank-1, respectively. These results indicate that single-prototype designs are insufficient for aerial-ground matching, and that decoupling view-specific prototypes mainly improves overall ranking quality and intra-view discrimination while remaining competitive in cross-view retrieval.

\begin{wraptable}{r}{0.55\textwidth}
	\centering
	\caption{\textbf{Computational complexity comparison.} Latency is measured as the average forward pass time over 200 runs.}
	\label{tab:complexity}
	\renewcommand{\arraystretch}{1.2}
	\resizebox{0.55\textwidth}{!}{%
		\begin{tabular}{c|ccccc}
			\hline
			\textbf{Setting} & Params (M) & GFLOPs & Latency (ms) & $\Delta$Params & $\Delta$Latency \\ \hline
            VDT     & 85.97  & 60.75 & 11.19 & -- & -- \\
            GSAlign & 111.67 & 76.92 & 11.44 & +29.9\% & +2.23\% \\ 
            \hline
            \rowcolor[gray]{0.92}\textbf{ProAlign} & 86.98 & 74.69 & 11.25 & +1.17\% & +0.54\% \\
			\hline
		\end{tabular}%
	}
\end{wraptable}

\noindent\textbf{Computational Complexity Analysis.} We analyze the computational efficiency of ProAlign in terms of trainable parameters, FLOPs, and inference latency, and compare it with VDT~\citep{zhang2024view} and GSAlign~\citep{gsalign} under the same input resolution ($256\times128$) and hardware configuration. ProAlign introduces additional parameters of 1.01M compared to VDT, which are contributed by the Layer-wise Progressive Positional Embedding (LPPE) module, including depthwise $3\times3$ convolutions, layer-specific positional embeddings, and learnable gating scalars. The Prototype Contrastive Learning (PCL) module does not introduce additional trainable parameters at inference time, as the prototype memory is maintained as non-trainable buffers used only during training. In terms of computational cost, ProAlign requires 74.69 GFLOPs per forward pass. Despite the increased computation, the inference latency remains comparable, increasing by only 0.06\,ms (0.54\%). In contrast, GSAlign introduces 25.7M additional parameters for TPS-based warping and increases latency by 2.23\%, whereas ProAlign achieves competitive alignment with substantially lower complexity and without relying on explicit geometric transformation.

\noindent\textbf{Spatial Attention Concentration.}\label{sec:drift}
We analyze layer-wise spatial attention concentration as an indirect
indicator of positional calibration. Lower normalized attention entropy
indicates that the global token attends to a smaller set of spatial
locations, but does not by itself establish correct spatial
correspondence. We therefore complement the quantitative analysis
with attention visualizations to examine whether the concentrated
responses fall on identity-relevant pedestrian regions. 
At layer $l$, we average the [CLS]-to-patch
attention weights over all heads and normalize
them to sum to one, obtaining a distribution
$\mathbf{a}^{(l)} \in \Delta^{N-1}$ over
the $N$ patch tokens ($N=128$ for
$256\times128$ inputs). Its normalized
entropy is
\begin{equation}
\mathcal{H}^{(l)} = -\frac{1}{\log N}\sum_{j=1}^{N} a^{(l)}_j \log a^{(l)}_j \in [0,1],
\end{equation}
where $\mathcal{H}^{(l)}{=}1$ means perfectly uniform attention (no spatial selectivity) and lower values indicate sharper focus. Intuitively, $N^{\mathcal{H}^{(l)}}$ can be read as the effective number of attended patches: $\mathcal{H}{=}1.0$ corresponds to all 128 patches, while $\mathcal{H}{=}0.81$ corresponds to about 51 patches. Table~\ref{Tbl:entropy} reports $\mathcal{H}^{(l)}$ averaged over paired aerial/ground test samples. The VDT baseline yields $\mathcal{H}^{(l)} \approx 1.000$ in all measured layers, i.e., its global token aggregates information almost uniformly, including background and occluded regions. In contrast, all LPPE variants reduce the entropy to 0.81--0.88 in Layers 7--10, where high-level semantics emerge, corresponding to attending to only about 40--55\% of the patches regardless of CPE depth. This agrees with the attention maps in Appendix~\ref{sup:vis} (Fig.~\ref{fig:attn_map}). Further layer-wise analysis is provided in Appendix~\ref{sup:drift}.

\begin{table}[!t]
\centering
\caption{\textbf{Normalized [CLS]-attention entropy $\mathcal{H}^{(l)}$ per layer} on CARGO. 1.0 denotes uniform attention over all 128 patches; lower values indicate sharper spatial focus. LPPE-$r$ applies CPE to the first $12r$ layers (LPPE-0.5 is the default ProAlign setting).}
\label{Tbl:entropy}
\renewcommand{\arraystretch}{1.2}
\resizebox{0.8\textwidth}{!}{%
\begin{tabular}{l|ccccccccc}
\hline
Layer & 3 & 4 & 5 & 6 & 7 & 8 & 9 & 10 & 11 \\ \hline
VDT & 1.000 & 1.000 & 1.000 & 1.000 & 1.000 & 1.000 & 1.000 & 1.000 & 1.000 \\
LPPE-0.25 & 0.982 & 0.982 & 0.977 & 0.951 & 0.881 & 0.851 & 0.864 & 0.857 & 0.986 \\
\rowcolor[gray]{0.92} LPPE-0.5  & 0.983 & 0.982 & 0.960 & 0.868 & \textbf{0.809} & \textbf{0.839} & \textbf{0.837} & 0.865 & 0.931 \\
LPPE-0.75 & 0.982 & 0.985 & 0.958 & \textbf{0.861} & 0.817 & 0.842 & 0.860 & 0.881 & 0.941 \\
LPPE-1.0  & 0.982 & 0.982 & 0.973 & 0.951 & 0.871 & 0.850 & 0.862 & \textbf{0.845} & 0.974 \\
\hline
\end{tabular}}
\end{table}

\begin{figure}[!t]
\vspace{-1em}
\centering
\includegraphics[width=0.9\linewidth]{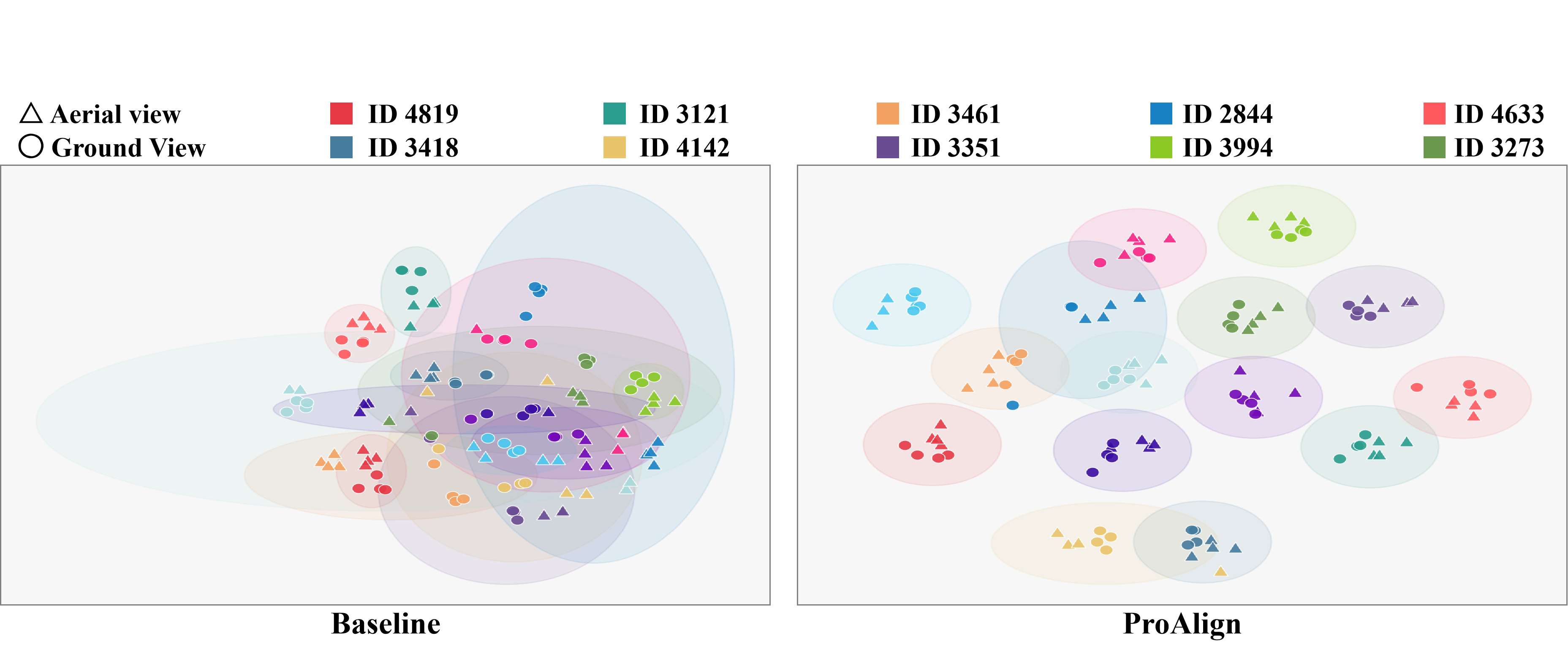}
\caption{\textbf{t-SNE visualization of feature distributions.}
$\triangle$ and $\bigcirc$ denote aerial and ground features, respectively. Baseline shows view-dependent clustering with fragmented identity distributions, while ProAlign produces more compact identity-wise clusters and reduces cross-view discrepancy.}
\label{fig:tsne_vis}
\vspace{-1em}
\end{figure}

\noindent\textbf{Feature Distribution Analysis.}
We further analyze the learned feature space to evaluate the effectiveness of the proposed PCL module. Following standard practice, we extract deep features from the test set and visualize their distributions using t-SNE, as shown in Fig.~\ref{fig:tsne_vis}. The baseline model exhibits a clear view-dependent clustering pattern, where samples are primarily grouped by viewpoint rather than identity. As a result, features of the same identity are scattered across different regions, indicating a pronounced cross-view semantic gap and insufficient sample-level alignment under large viewpoint variations. In contrast, ProAlign produces a more structured embedding space. Features become significantly more identity-consistent, with aerial and ground samples of the same person forming compact and unified clusters. Meanwhile, inter-identity boundaries become more distinct, suggesting improved discriminability. These observations suggest that the proposed framework improves cross-view feature consistency and identity discrimination.

\section{Conclusion}

In this paper, we propose ProAlign for aerial-ground person re-identification, which addresses positional drift and semantic inconsistency in cross-view scenarios. By introducing a Layer-wise Progressive Positional Embedding (LPPE) and a View-Aware Prototype Contrastive Learning (PCL), the model performs depth-adaptive positional calibration and population-level semantic alignment within a unified framework. Extensive experiments on the CARGO dataset demonstrate state-of-the-art performance across multiple protocols. We further evaluate generalization on the AG-ReID and AG-ReID v2 datasets, where consistent improvements are observed, validating the robustness of our approach under real-world cross-view variations and domain gaps.

\section{Limitation and Broader Impact}
\label{sec:limitation}
Despite achieving state-of-the-art performance in AG-ReID, ProAlign still has several limitations that merit further investigation. First, the Layer-wise Progressive Positional Embedding (LPPE) relies on a predefined partition between conditional and static positional encodings across network depth. Such a fixed design lacks instance-level adaptivity and may be sub-optimal under varying degrees of spatial deformation across different samples.
Second, the View-Aware Prototype Contrastive Learning (PCL) module updates prototypes using mini-batch statistics. As a result, its cross-view alignment is inherently constrained by batch composition; when an identity does not have paired samples across views within a batch, the corresponding contrastive supervision cannot be fully exploited. Future work will explore more adaptive designs, including instance-aware dynamic gating for positional calibration and cross-batch memory mechanisms to decouple prototype learning from individual batch constraints, thereby improving robustness in more complex and large-scale real-world scenarios.

\paragraph{Acknowledgements.}
This research was supported in part by the National Natural Science Foundation of China under grants  No. 62625207, 62602479, 62302271,  the National Cryptologic Science Fund of China under grant No. 2025NCSF02027, and the China Postdoctoral Science Foundation - Hubei Joint Support Program under Grant Number 2026T077HB.
\medskip
\small
\bibliographystyle{unsrt}
\bibliography{neurips_2025}

\newpage
\appendix

\section{Supplementary Material}
\subsection{Visualization Analysis}
\label{sup:vis}

\begin{figure}[!h]
\centering
\includegraphics[width=0.85\linewidth]{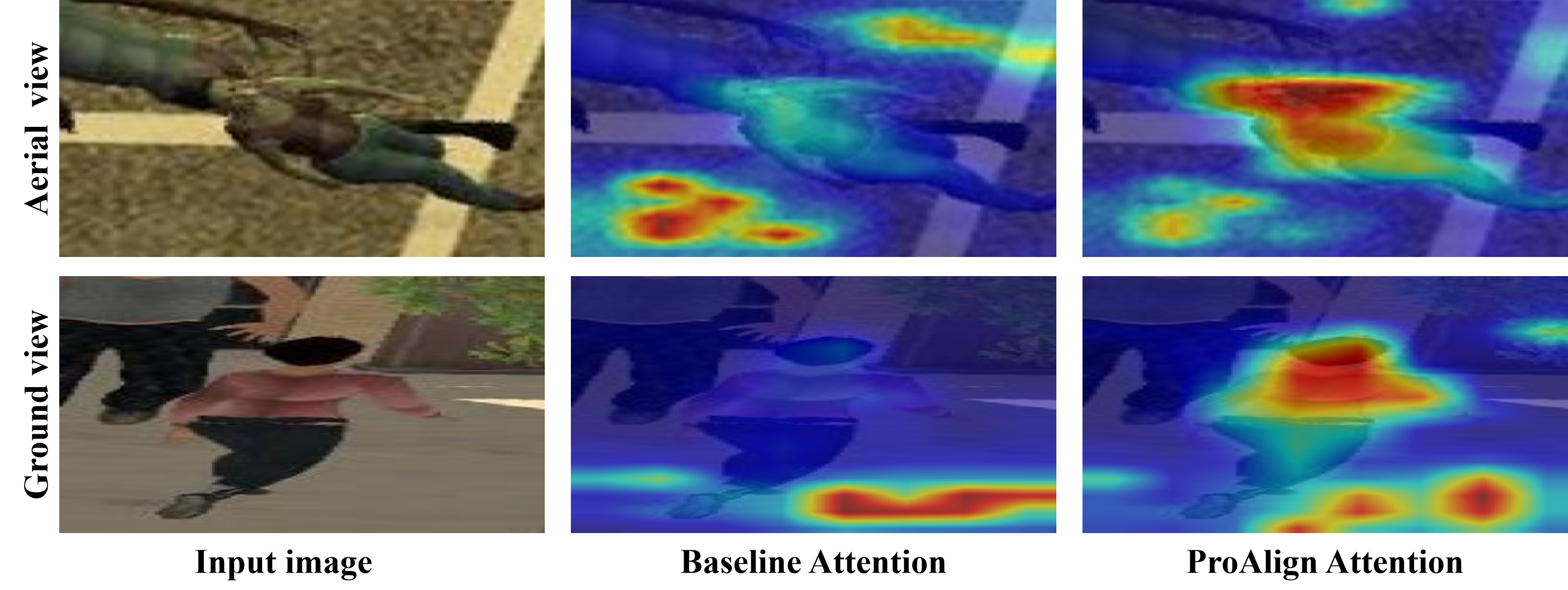}
\caption{\textbf{Attention map comparison between baseline and ProAlign.} In both aerial and ground views, the baseline attention is scattered due to severe viewpoint distortion, while LPPE produces more focused responses on the person.}
\label{fig:attn_map}
\vspace{-1em}
\end{figure}

\noindent\textbf{Effectiveness Analysis of the LPPE Module.}
We visualize attention maps to analyze the effect of the Layer-wise Progressive Positional Embedding (LPPE) module by comparing the baseline VDT with its LPPE-enhanced variant. As shown in Fig.~\ref{fig:attn_map}, the baseline model exhibits scattered attention and frequently focuses on background regions, indicating severe spatial misalignment under large viewpoint changes. In contrast, ProAlign produces more concentrated and accurate attention distributions, consistently focusing on the target person under both aerial and ground views. This suggests that LPPE effectively improves spatial correspondence by progressively correcting positional deviations across transformer layers, leading to more stable and reliable feature localization in cross-view scenarios.

\subsection{Additional Comparisons with State-of-the-Arts}
\label{sec:agreid}

\begin{wraptable}{r}{0.55\textwidth}
	\centering
	\fontsize{8.5pt}{\baselineskip}\selectfont 
	\renewcommand\tabcolsep{3pt} 
	\renewcommand\arraystretch{1.1} 
\caption{\textbf{Performance comparison of the mainstream methods under two settings of the AG-ReID dataset.} Results are reported under matching protocols. The best values per column are shown in \textbf{bold}.}
\begin{tabular}{c|ccc|ccc}
			\hline
			\multirow{2}{*}{\textbf{Setting}} & \multicolumn{3}{c|}{Protocol 1: A$\leftrightarrow$G}  & \multicolumn{3}{c}{Protocol 2: G$\leftrightarrow$A} \\ \cline{2-7}
			& Rank1 & mAP & mINP & Rank1 & mAP & mINP \\
\hline
SBS~\cite{he2023fastreid} & 73.54 & 59.77 & - & 73.70 & 62.27 & - \\
BoT~\cite{luo2019bag}& 70.01 & 55.47 & - & 71.20 & 58.83 & - \\
OSNet~\cite{zhou2021learning} & 72.59 & 58.32 & - & 74.22 & 60.99 & - \\
VV ~\cite{kuma2019vehicle, kumar2020strong} & 77.22 & 67.23 & 41.43 & 79.73 & 69.83 & 42.37 \\
Explain~\cite{nguyen2023aerial} & 81.47 & 72.61 & - & 82.85 & 73.39 & - \\
VDT~\cite{zhang2024view} & 82.91 & 74.44 & 51.06 & 86.59 & 78.57 & 52.87 \\
VIF~\cite{Khalid_2025_ICCV}	& 83.75 & 75.22 & 51.57 & \textbf{87.32} & \textbf{79.19} & 52.98 \\
\hline
\rowcolor[gray]{0.92} \textbf{ProAlign} & \textbf{86.18} & \textbf{77.16} & \textbf{54.06} & 85.86 & 78.94 & \textbf{53.38} \\
 \hline
		\end{tabular}%
\vspace{-3mm}
	\label{Tbl:agreid}
\end{wraptable}

\textbf{Results on AG-ReID.}
We first evaluate ProAlign on the real-world AG-ReID dataset to assess its generalization ability. As shown in Table~\ref{Tbl:agreid}, ProAlign achieves state-of-the-art performance under the A$\leftrightarrow$G setting, reaching 86.18\% Rank-1 and 77.16\% mAP, outperforming strong baselines across both metrics. For the G$\leftrightarrow$A setting, VIF achieves slightly higher performance with 87.32\% Rank-1 and 79.19\% mAP, whereas ProAlign still maintains competitive results and achieves the best mINP with a clear margin, improving robustness under cross-view variations. Overall, these results demonstrate that ProAlign provides balanced improvements across different matching directions while preserving stable performance in real-world conditions.

\noindent\textbf{Results on AG-ReID v2.}
We further evaluate ProAlign on the more challenging AG-ReID v2 dataset. As shown in Table~\ref{Tbl:agreidv2}, ProAlign achieves the highest Rank-1 accuracy of 91.68\% under the A$\leftrightarrow$G protocol, improving upon the strongest competing result by 2.91 \%. Although its Rank-1 performance is not consistently the best across all directions, ProAlign achieves the highest mAP under the G$\leftrightarrow$A, A$\leftrightarrow$W, and W$\leftrightarrow$A protocols, outperforming the strongest competing results by 7.51\%, 6.65\%, and 7.65 \%, respectively. These results demonstrate that ProAlign effectively improves overall retrieval quality across diverse real-world matching scenarios.

\begin{table*}[!t]
	\centering
	\caption{\textbf{Performance comparison of the mainstream methods under four settings of the AG-ReID v2 dataset. } Results are reported under four matching protocols. The best values per column are shown in \textbf{bold}.}
	\label{Tbl:agreidv2}
	\renewcommand{\arraystretch}{1.2}
	\resizebox{0.9\textwidth}{!}{%
		\begin{tabular}{c|cc|cc|cc|cc}
			\hline
			\multirow{2}{*}{\textbf{Setting}} & \multicolumn{2}{c|}{Protocol 1: A$\leftrightarrow$G} & \multicolumn{2}{c|}{Protocol 2: G$\leftrightarrow$A} & \multicolumn{2}{c|}{Protocol 3: A$\leftrightarrow$W} & \multicolumn{2}{c}{Protocol 4: W$\leftrightarrow$A} \\ \cline{2-9}
			& Rank1 & mAP & Rank1 & mAP & Rank1 & mAP  & Rank1 & mAP \\
 \hline
 BoT~\cite{luo2019bag} & 85.40 & 77.03 & 84.65 & 75.90 & 89.77 & 80.48 & 84.65 & 76.90 \\
Explain~\cite{nguyen2023aerial}& 87.70 & 79.00 & 87.35 & 78.24 & \textbf{93.67} & 83.14 & 87.73 & 79.08 \\
 VDT~\cite{zhang2024view} & 86.46 & 79.13 & 86.14 & 78.12 & 90.00 & 82.21 & 85.26 & 78.52 \\
AG-ReIDv2~\cite{nguyen2024ag} & 88.77 & 80.72 & 87.86 & 78.51 & 93.62 & 84.85 & \textbf{88.61} & 80.11 \\
SeCap~\cite{wang2025secap} & 88.12 & \textbf{80.84} & \textbf{88.24} & 79.99 & 91.44 & 84.01 & 87.56 & 80.15 \\
\hline
\rowcolor[gray]{0.92}
\textbf{ProAlign}  &\textbf{91.68} & 80.17 & 87.19 & \textbf{87.50} & 91.58 & \textbf{91.50} & 87.09 & \textbf{87.80} 
 \\
\hline
		\end{tabular}%
	}
\end{table*}

\subsection{Sensitivity Analysis}
\noindent\textbf{Momentum coefficient for prototype update.}
We study the sensitivity of the momentum coefficient $m$ used in the prototype update of the PCL memory bank, which controls how rapidly prototypes adapt to new mini-batches. As shown in Table~\ref{Tbl:momentum}, we evaluate $m \in \{0.1, 0.2, 0.3\}$. Overall, $m=0.2$ achieves the strongest
overall retrieval performance, while
$m=0.1$ yields better results under
the A$\leftrightarrow$G protocol. A smaller momentum $m=0.1$ leads to slower prototype updates, which benefits the A$\leftrightarrow$G setting but degrades global performance in mAP and mINP. In contrast, a larger momentum $m=0.3$ makes prototypes more sensitive to batch noise, resulting in performance degradation, particularly in the ALL setting. These results indicate that a moderate momentum effectively balances stability and adaptability, enabling robust prototype estimation under noisy and imbalanced cross-view data.

\begin{table*}[!t]
	\centering
	\caption{\textbf{Impact of the momentum coefficient ($m$) for prototype update in PCL.} Results are reported under four matching protocols on the CARGO dataset. The best values per column are shown in \textbf{bold}.}
	\label{Tbl:momentum}
	\renewcommand{\arraystretch}{1.2}
	\resizebox{\textwidth}{!}{%
		\begin{tabular}{c|ccc|ccc|ccc|ccc}
			\hline
			\multirow{2}{*}{\textbf{Setting}} & \multicolumn{3}{c|}{Protocol 1: ALL} & \multicolumn{3}{c|}{Protocol 2: G$\leftrightarrow$G} & \multicolumn{3}{c|}{Protocol 3: A$\leftrightarrow$A} & \multicolumn{3}{c}{Protocol 4: A$\leftrightarrow$G} \\ \cline{2-13} 
			& Rank1 & mAP & mINP & Rank1 & mAP & mINP & Rank1 & mAP & mINP & Rank1 & mAP & mINP \\ 
            \hline
            $m=0.1$ & 67.63 & 60.75 & 47.97 & 81.25 & 75.57 & 66.16 & 80.00 & 67.50 & 51.37 & \textbf{67.02} & \textbf{62.61} & \textbf{52.69} \\ 
            $m=0.2$ & \textbf{68.27} & \textbf{62.47} & \textbf{51.39} & \textbf{84.82} & \textbf{80.22} & \textbf{72.51} & \textbf{85.00} & \textbf{70.05} & \textbf{54.66} & 60.64 & 58.44 & 49.86 \\ 
            $m=0.3$ & 67.31 & 59.24 & 45.95 & 83.93 & 75.79 & 65.32 & 82.50 & 67.00 & 49.45 & 64.89 & 61.42 & 51.48 \\ 
            \hline
		\end{tabular}%
	}\vspace{-1em}
\end{table*}

\subsection{Further Analysis of Spatial Attention}
\label{sup:drift}

This section extends the entropy analysis in Sec.~\ref{sec:drift}. As shown in Table~\ref{Tbl:entropy}, all LPPE variants follow a consistent layer-wise pattern. In the shallow layers (3--5), the entropy remains close to uniform ($\geq$0.958) for all variants, which is expected since these layers mainly encode low-level texture rather than semantic body parts. The entropy then drops markedly in the middle layers (7--10), where semantic part representations emerge and spatial correspondence matters most, and rises again in Layer 11 (0.93--0.99), suggesting renewed attention to broader contextual information in deeper representations. The VDT baseline, by contrast, remains near-uniform in all layers, indicating that without positional calibration its global token never recovers spatial selectivity. This spatial focus is also reflected in retrieval performance: LPPE-0.5, the configuration adopted in ProAlign, attains the lowest entropy in Layers 7--9, which coincides with its best ALL mAP/mINP in Table~\ref{Tbl:cpe_ratio}, suggesting that sharper focus in the middle layers translates into better overall ranking quality.

These results also explain why LPPE calibrates only the patch tokens. The class and view tokens have no spatial coordinates and therefore cannot drift in the same sense as patch tokens; they access spatial information only through attention. Once LPPE calibrates the patch tokens, the global tokens attend to more informative regions without direct modification, as reflected by the reduced [CLS] entropy. Finally, we note that lower entropy measures how concentrated the attention is rather than whether it falls on the correct region. We therefore complement it with the attention visualizations in Fig.~\ref{fig:attn_map}, which confirm that the concentrated regions lie on the person rather than the background.



\end{document}